\pdfoutput=1
\documentclass[11pt]{article}

\usepackage[final]{acl}

\usepackage{times}
\usepackage{latexsym}

\usepackage[T1]{fontenc}
\usepackage[utf8]{inputenc}

\usepackage{microtype}

\usepackage{inconsolata}

\usepackage{graphicx}
\usepackage{graphicx}
\usepackage{booktabs}
\usepackage{pifont}
\usepackage{amsthm,amsmath,amssymb}
\usepackage{multirow} 
\usepackage{tcolorbox}
\usepackage{hyperref}
\usepackage{enumitem}
\usepackage{multicol}
\usepackage{booktabs} 
\usepackage{fontawesome}
\usepackage{array}
\usepackage{colortbl} % 用于设置表格颜色
\usepackage{xcolor} % 提供颜色支持
\usepackage{subcaption}
\newcommand{\up}[1]{\textsubscript{\color{green!70!black}{↑#1}}}

\title{TinyJudge: Unverifiable Constraint Alignment via Lightweight Specialist Ensembles}
\author{
 \textbf{Yirong Zeng\textsuperscript{1\thanks{Authors contributed equally, \\$\dagger$Correspondence: \{xding,bbcai\}@ir.hit.edu.cn}}},
 \textbf{Yufei Liu\textsuperscript{2*}},
 \textbf{Xiao Ding\textsuperscript{1$\dagger$}},
  \textbf{Yutai Hou\textsuperscript{3}},  
 \textbf{Yuxian Wang\textsuperscript{3}}, \\
 \textbf{Wu Ning\textsuperscript{3}},
 \textbf{Haonan Song\textsuperscript{3}},
 \textbf{Qixun Zhang\textsuperscript{2}},
 \textbf{Yuxiang He},
 \textbf{Dandan Tu\textsuperscript{3}},
 \textbf{Bibo Cai\textsuperscript{1$\dagger$}},
 \textbf{Ting Liu\textsuperscript{1}},
\\
 \textsuperscript{1}Harbin Institute of Technology SCIR Lab,
 \textsuperscript{2}Peking University, \\
 \textsuperscript{3}Huawei Technologies Co., Ltd,
\\
 \small{ 
  Work done while interning at Huawei.
 }
}

\begin{document}
\maketitle
\begin{abstract}
Instruction Following (IF) is a core capability of LLMs, requiring strict adherence to diverse constraints, ranging from verifiable ones (e.g., output length) to unverifiable ones (e.g., tone).
Reinforcement learning with verifiable rewards has emerged as a paradigm for IF tasks, leveraging LLM-as-a-judge to assess unverifiable constraints.
However, we empirically find that this approach remains a significant bottleneck, suffering from severe reward hacking and higher computational overhead.
In this work, we first analyze the generalization capabilities of unverifiable constraints and discover that specific constraints exhibit distinct, high-generalization patterns. 
Motivated by this, we propose TinyJudge, a framework that employs an ensemble of specialized tiny language models ($\sim0.6B$) to provide rewards for soft constraints. 
By distilling expertise from frontier models into these tiny models, it achieves high-precision, lightweight evaluation.
Extensive evaluations across five benchmarks demonstrate that TinyJudge outperforms the baselines by $\sim10\%$ in average performance and $12\%$ in reward precision. 
Crucially, it also achieves a $3\times$ speedup in total training time.
Our work provides a scalable and robust path for aligning LLMs with unverifiable human instructions.

\end{abstract}

\section{Introduction}
IF evaluates a model's ability to adhere to specific constraints, typically categorized into hard constraints (verifiable via rigid rules, e.g., length limits) and soft constraints (unverifiable by rules and requiring semantic judgment, e.g., tone or style)~\citep{tam2024let,zhou2023instruction}.
% \footnote{In this paper, \textit{soft} and \textit{semantic} constraint are used interchangeably, as are \textit{hard} and \textit{rule} constraint.}
Current research is increasingly shifting focus from hard-only constraints toward scaling diverse constraints to enhance model generalization across real-world scenarios~\citep{jiang2024followbench,zhang2025cfbench}.

\begin{figure}[t]
    \centering
  \includegraphics[width=\columnwidth]{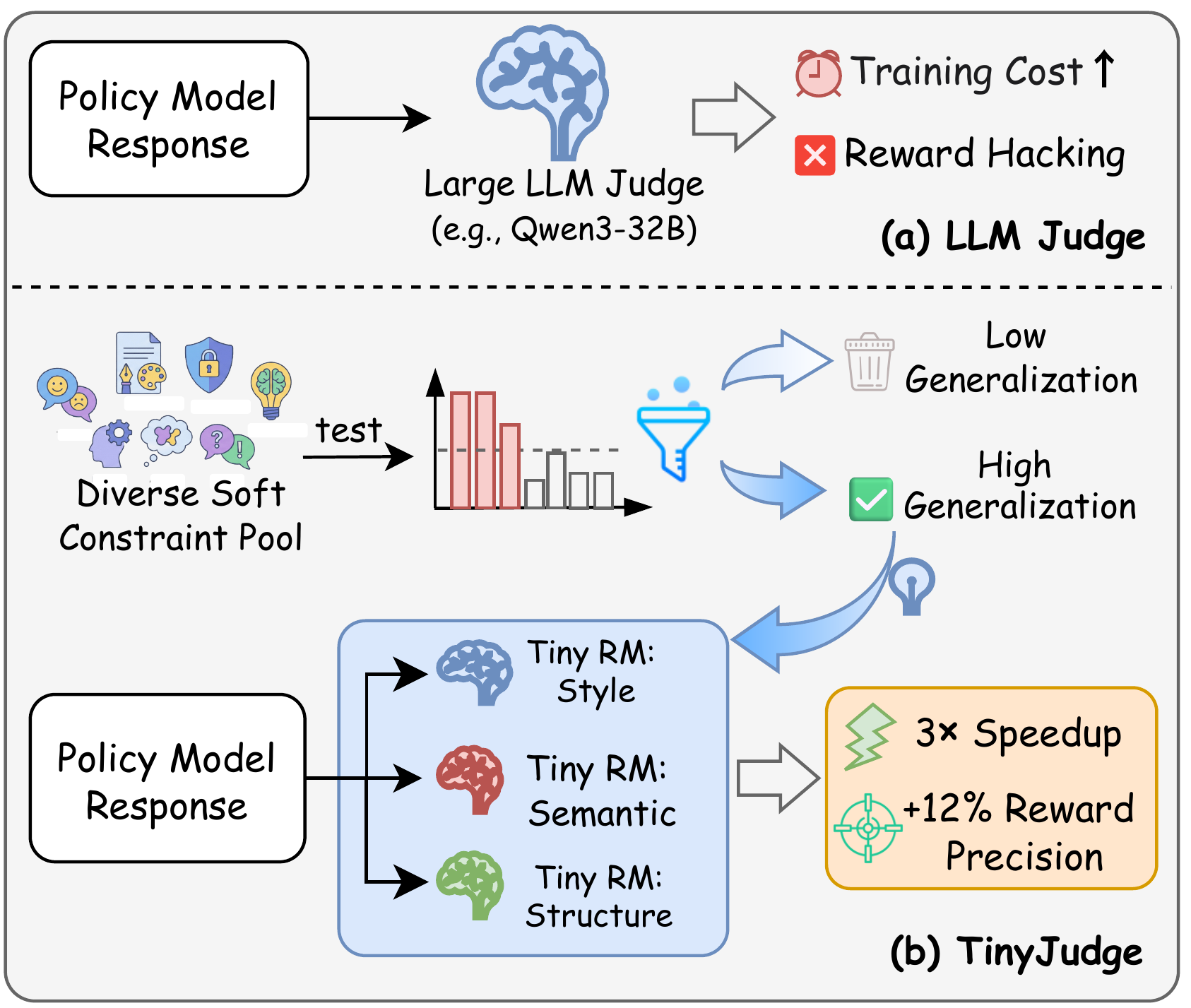}
  \caption{
    The challenges of current LLM-based reward in handling unverifiable soft constraints and our solution, TinyJudge.
    (a) Existing LLM-as-a-judge (\textbf{Top}): suffers from high computational overhead and reward hacking. 
    (b) Motivated by insights from generalization analysis (\textbf{Middle}), we employ a limited number of tiny expert-level models as reward, achieving superior reward precision and speedup in judging (\textbf{Bottom}).
    }
  \label{fig:intro}
\end{figure}

Recently, Reinforcement Learning from Verifiable Rewards (RLVR), driving the development of advanced models~\citep{guo2025deepseek,OpenAIGPT5}, has emerged as a dominant paradigm for enhancing instruction-following capabilities.
To further this progress, the research community is now scaling RLVR by integrating heterogeneous constraints to achieve broad generalization~\citep{ren2025instructions,peng2025verif}.
This strategy combines rule-based rewards for verifiable hard constraints with LLM-as-a-judge for unverifiable soft constraints~\citep{guo2025recast,pyatkin2025generalizing}. 
% The research community considers LLM-as-a-judge a viable and practical reward mechanism, yet its underlying effectiveness remains underexamined.
However, this prevailing strategy relies heavily on the assumption that LLM-as-a-judge acts as a reliable evaluator. 
This raises a fundamental question: Is the current LLM-as-a-judge paradigm truly effective and robust enough to serve as a precise reward signal for RL training?

To address this question, we investigate the effectiveness of LLM-as-a-judge for unverifiable constraints in Section~\S\ref{sec:pilot}.
Our empirical analysis reveals two critical issues:
(1) {Severe reward bias:} When evaluating outputs against multiple constraints simultaneously, the LLM judge exhibits a significant tendency to overlook violations (i.e., failing to penalize errors), resulting in low reward precision.
(2) {Higher training overhead:} Directly employing a frontier LLM as the reward model incurs a 3$\times$ increase in training time overhead.
As illustrated in Figure~\ref{fig:intro}, these issues severely restrict both the practicality and the performance ceiling of the method.
{Consequently, addressing these inefficiencies is a prerequisite for scaling RLVR to diverse, real-world IF tasks.}

To this end, we first analyze the generalization capabilities of various unverifiable constraints. 
It reveals that some specific constraints exhibit significantly higher generalization than others.
This suggests that decoupling evaluation, by assessing specific constraints individually rather than via a monolithic model for all constraints simultaneously, can mitigate reward bias.
Motivated by this insight, we propose \textit{TinyJudge}, a framework that employs tiny language models (e.g., 0.6B) to provide rewards exclusively for these high-generalization constraints. 
It distills expertise from frontier models into a specialized tiny model only for each high-generalization constraint.
% 放弃使用其他软约束类型
During RLVR training, reward signals are generated through a hybrid system: each unverifiable soft constraint is evaluated by the corresponding tiny expert model, while verifiable hard constraints are processed via code-based rules. 
This ensemble reward module provides high-precision, low-latency feedback to the policy model.

We evaluated our method across five IF benchmarks (e.g., IFEval and CFBench). 
Under identical experimental settings, TinyJudge achieves a $\sim$ 10\% performance gain over the base model, while significantly outperforming other established baselines.
The efficiency of our approach is equally notable. 
Compared to previous LLM-based reward systems, our method improves reward precision by 12\%. 
Furthermore, TinyJudge achieves a 6$\times$ speedup in judging latency per response, resulting in a 3$\times$ reduction in total training time. 
Notably, its computational overhead is nearly identical to that of training with only hard constraints, demonstrating its extreme efficiency.
In summary, this work paves the way for efficient and robust alignment with complex unverifiable human instructions.

% Our contributions are threefold:
% \begin{enumerate}
%     \item Problem: identifying a counter-intuitive phenomenon, and reveal the distinct roles of reward precision and constraint diversity.
%     % in RLVR.  demonstrating the critical importance of the former.
%      % yield a general constraint operator that 
%     \item Method: proposing a simple yet effective data-centric refinement strategy to enable high-precision proxy training.
% \end{enumerate}

\section{Preliminaries}
\subsection{Task Definition: Instruction Following}
Let $\mathcal{I}$ denote the space of natural language instructions and $\mathcal{Y}$ denote the space of model-generated responses. An instruction $I \in \mathcal{I}$ typically contains a core task accompanied by a set of $n$ constraints $\mathcal{C} = \{c_1, c_2, \dots, c_n\}$. We categorize these constraints into two disjoint subsets:

\begin{itemize}
    \item {Hard Constraints:} Objective requirements that can be verified via deterministic programs or rules (e.g., \textit{"output in JSON format"} ).
    \item {Soft Constraints:} Subjective qualities that require semantic understanding to evaluate (e.g., "\textit{maintain a professional tone}").
\end{itemize}

The goal of LLM instruction following is to learn a mapping function $f: \mathcal{I} \to \mathcal{Y}$, where $\mathcal{Y}$ represents the space of target responses that satisfy both the semantic intent and the explicit constraints defined in $\mathcal{I}$.
Formally, given an instruction $I \in \mathcal{I}$, an LLM parameterized by $\theta$ generates a response $\hat{y}$ by modeling the conditional probability:
\begin{equation}
    \small
    P(\hat{y} | I; \theta) = \prod_{t=1}^{T} P(y_t | I, y_{<t}; \theta),
\end{equation}
where $T$ is the sequence length. 
The task objective is to minimize the discrepancy between the generated response $\hat{y}$ and the optimal response $y^*$.

\begin{figure*}[th]
    \small
    \centering
    \includegraphics[width=0.85\linewidth]{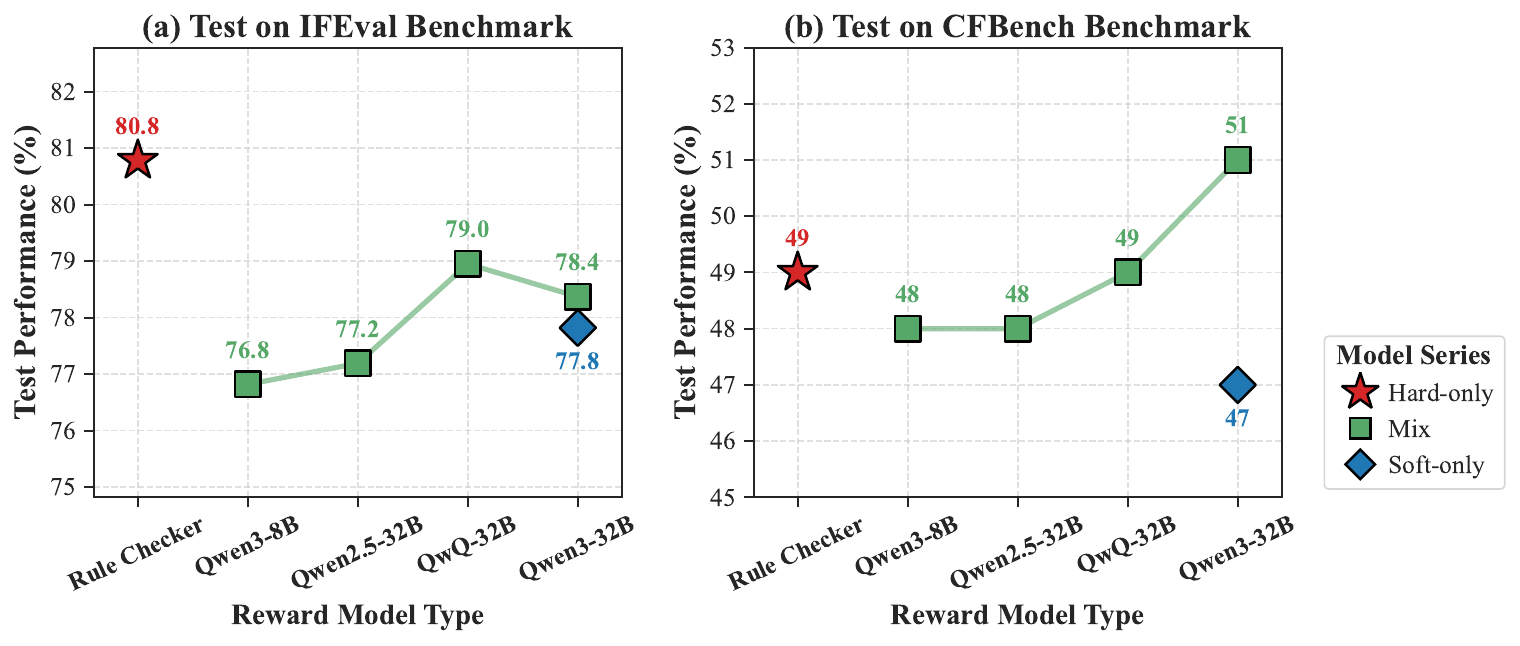} 
    \caption{
      Test performance of the model under different reward models and training data configurations on IFEval (evaluating hard constraints) and CFBench (evaluating mixed constraints). 
      Results show that: (1) the \textit{hard-only} model outperforms the \textit{soft-only} variant, and also performs comparably to the mixed-constraint model; 
      and (2) a stronger reward model leads to better test performance.
    }
  \label{fig:pre_study1}
\end{figure*}

\subsection{GRPO as an RLVR algorithm}
\label{sec:grpo}
We employ Group Relative Policy Optimization (GRPO)~\citep{shao2024deepseekmath} as our core RL algorithm. 
Unlike traditional PPO~\citep{schulman2017proximal,engstrom2020implementation} which relies on a value function critic, GRPO leverages group-based sampling to estimate baselines. This approach generates multiple candidate responses for the same instruction and computes advantage estimates through intra-group comparisons, effectively capturing the relative quality differences among responses.

Formally, for each input instruction $x$, the policy $\pi_\theta$ samples a group of $G$ candidate responses $\{y_i\}_{i=1}^G$. The optimization objective is defined as:
\begin{equation}
    \small
    \begin{split}
    \mathcal{J}(\theta) & =  \mathbb{E}_{\substack{x \sim P(X), \\ \{y_i\}_{i=1}^G \sim \pi_{\theta_{\text{old}}}(Y|x)}} \Bigg[ \frac{1}{G} \sum_{i=1}^G  \frac{1}{|y_i|} \sum_{t=1}^{|y_i|} \bigg\{  \\
    & \min \left( \rho_{i,t} \hat{A}_{i,t}, \text{clip} \left( * \right) \hat{A}_{i,t} \right) - \beta \mathbb{D}_{\text{KL}} \left[ \pi_\theta \| \pi_{\text{ref}} \right] \bigg\} \Bigg],
    \end{split}
\end{equation}
where the \(\text{clip}(*) \) denotes \( \text{clip}\left( \rho_{i,t}, 1-\varepsilon, 1+\varepsilon \right) \).
Note that $\varepsilon$ is a small constant for numerical stability.
The advantage $\hat{A}_i$ is standardized within the group to reduce variance:
\begin{equation}
\small
\begin{aligned}
\mu &= \frac{1}{G} \sum_{i=1}^G r_i, & 
\sigma &= \sqrt{\frac{1}{G} \sum_{i=1}^G (r_i - \mu)^2 + \epsilon} \\
\hat{A}_i &= \frac{r_i - \mu}{\sigma}, & 
\rho_{i,t} &= \frac{\pi_\theta(y_{i,t} \mid x, y_{i,<t})}{\pi_{\theta_{\text{old}}}(y_{i,t} \mid x, y_{i,<t})},
\end{aligned}
\end{equation}
where \( r_i\) denotes the reward signal assigned by the reward model to the \(i\)-th response.

\section{Pilot Experiments}
\label{sec:pilot}
In this section, we thoroughly evaluate the effectiveness of LLM-as-a-judge and present a key insight to improve performance.

\subsection{Examining LLM-as-a-Judge}
% To this end, we conduct a comparative analysis using various LLMs as automated judges across multiple experimental settings. 

\begin{figure}[t]
    \centering
  \includegraphics[width=\columnwidth]{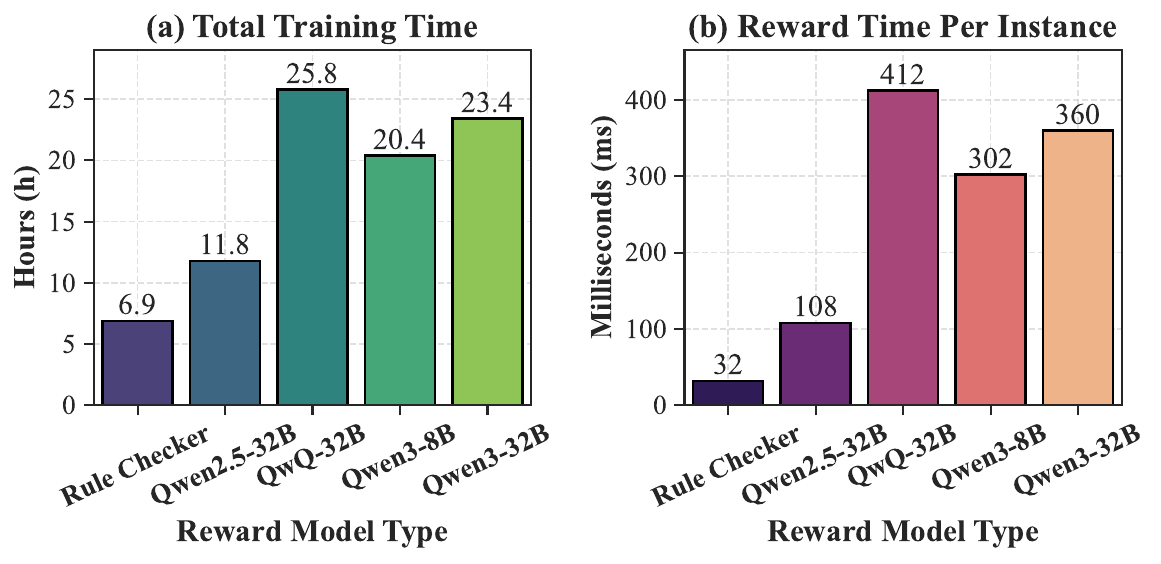}
  \caption{Comparison of training efficiency. (a) total training time (up to 100 iteration steps), (b) average reward latency per instance. 
  The LLM-based approach significantly increases computational costs compared to rule-based rewards.
    }
  \label{fig:time_cost}
\end{figure}

\paragraph{Experimental Setup.}
\label{sec:traindata}
For model training, we utilize the \textit{VerInstruct} dataset~\citep{peng2025verif}, comprising 22,000 instances. The dataset is partitioned into two constraint types: {soft constraints} (77.7\%), which are evaluated via LLM-as-a-judge, and {hard constraints} (22.3\%), which are validated through deterministic code-based rules. 
To investigate the influence of data distribution, we categorize the training data into three subsets: \textit{hard-only}, \textit{soft-only}, and the original \textit{mix} set.
We employ Qwen2.5-7B-Instruct~\citep{qwen2025qwen25} as the base model. 
To analyze the scaling and reasoning capabilities of reward models, we select a diverse suite of LLMs to serve as judges, specifically: Qwen2.5-32B, QwQ-32B~\citep{qwq32b}, Qwen3-8B~\citep{yang2025qwen3}, and Qwen3-32B~\citep{yang2025qwen3} \footnote{The closed-source LLMs accessing via APIs are impractical as reward models in RLVR training due to API rate limits.}. 

\paragraph{Test Performance.} 
We train models separately on the \textit{hard-only}, \textit{soft-only}, and \textit{mix} datasets and evaluate them on the hard constraint benchmark IFEval~\citep{zhou2023instruction} and mixed constraint benchmark CFBench~\citep{zhang2025cfbench}, details in Section \S\ref{sec:setup}. 
As illustrated in Figure \ref{fig:pre_study1}, the model trained on \textit{hard-only} data consistently outperforms the \textit{soft-only} and \textit{mix} variants (i.e., including both hard and soft constraint).
In particular, the hard-only model outperforms the \textit{soft-only} model (judged by Qwen3-32B) by 3.0\% and the \textit{mix} model by 2.4\% on IFEval.
This finding also holds on the out-of-distribution (OOD) benchmark CFBench.
This observation suggests that using an LLM as a judge for soft constraints does not confer broader OOD generalization to the model.
This suggests that the LLM judge introduces undesirable biases, highlighting that naively employing an LLM as a reward for unverifiable soft constraints is imperfect and potentially harmful.

\paragraph{Time Efficiency.}
We evaluate the computational overhead in terms of total training time and per-reward latency under identical experimental settings. 
Specifically, we compare the \textit{hard-only} model using rule-based rewards against the \textit{mix} model that combines rule-based rewards (for hard constraints) and LLM-based rewards (Qwen3-32B for soft constraints).
For reward latency of each instance, we report the average time required to judge a response across 1,000 randomly selected responses. 
As shown in Figure \ref{fig:time_cost}, employing an LLM-as-a-judge (e.g., Qwen3-32B) increases the total training time by $\sim339\%$ and the reward latency of each instance by 11$\times$ compared to the baseline. 
In contrast, rule-based rewards incur negligible overhead ($\sim$30 ms per instance). 
These results highlight the significant computational bottleneck introduced by LLM-based evaluation in the training loop.

\paragraph{Cross-Model Validation.} 
To ensure the validity of the above observation, we scale our experiments across different model architectures (e.g., Qwen2.5~\citep{qwen2025qwen25}, Qwen3~\citep{yang2025qwen3}, LLama3.2~\citep{dubey2024llama}) and sizes (3B, 7B, 8B, 32B). 
We also evaluate the model on three additional benchmarks (see Section \S\ref{sec:setup}).
The results are shown in Appendix \S \ref{sec:exp_details} Table \ref{tab:robust}.
The {soft-constraint models} remain the worst across all settings, suggesting that this is a fundamental limitation of the LLM-as-a-judge paradigm rather than a model-specific artifact.

\begin{figure}[t]
    \centering
  \includegraphics[width=0.85\columnwidth]{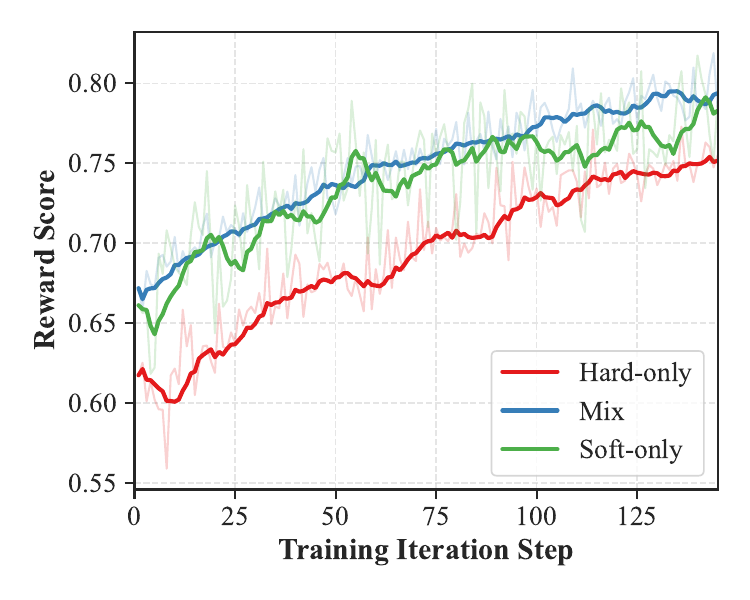}
  \caption{The training curves of models trained in hard-only, soft-only and mixed constraint data, respectively. 
  The \textit{Soft-only} model achieves higher reward scores than \textit{Hard-only} model.
  }
  \label{fig:training_curves}
\end{figure}

\paragraph{Visualization of Training.} 
To further understand the above observation, we visualize the model training curves (Qwen3-32B as the reward for soft constraint) in Figure~\ref{fig:training_curves}.
We surprisingly find that soft-only models achieve significantly higher reward scores during training yet ultimately deliver lower test performance.
This is mainly because soft-only models explore biases in LLM-as-a-judge to game the reward scores, rather than genuinely mastering constraint adherence. 
% In contrast, rewards derived from code-based rules provide a more faithful and reliable measure of actual model capability.
This finding indicates that LLM-based rewards lead to inflated reward scores, which do not translate into improved test performance.
Therefore, we argue that directly using LLMs as rewards is currently unreliable.

% \begin{figure}[th]
%     \small
%     \centering
%     \begin{subfigure}[t]{0.5\linewidth}
%         \includegraphics[width=0.99\linewidth]{./llm_reward_precision_a.pdf} 
%         % \caption{Evaluation of reward reliability on IFEval}
%         \label{fig:exp1a}
%     \end{subfigure}
%     % \hfill
%     \begin{subfigure}[t]{0.49\linewidth}
%         \includegraphics[width=0.99\linewidth]{./llm_reward_precision_b.pdf}
%         % \caption{Evaluation of reward reliability on CFBench.}
%         \label{fig:exp1b}
%     \end{subfigure}
%   \caption{Evaluation of reward reliability on hard constraint (a) and  soft constraint (b).
%   }
%   \label{fig:reward}
% \end{figure}

% \begin{figure}[th]
%     \small
%     \centering
%     \includegraphics[width=\linewidth]{./llm_reward_combined.pdf}
%     \caption{Evaluation of reward reliability on hard constraint (a) and  soft constraint (b).
%   }
%   \label{fig:llm_reward}
% \end{figure}

\begin{table}[t]
    \centering
    \small
    \begin{tabular}{l|c|c}
        \toprule
        \multirow{2}{*}{\textbf{Reward Model}} & \multicolumn{1}{c|}{\cellcolor{lightgray!40}{Hard Constraint}} & \multicolumn{1}{c}{\cellcolor{lightgray!40}{Soft Constraint}} \\
        & \textbf{Precision} & \textbf{Precision} \\
        \midrule
        Rule Checker & \textbf{96.0} & -- \\ \midrule
        
        \multicolumn{3}{c}{\cellcolor{lightgray!40}{Batch Judgment}} \\ \midrule
        Gemini-2.5-pro & 86.0 & \textbf{86.3} \\
        Qwen-3-32B & 76.5\textsubscript{\textcolor{red!70!black}{↓19.5}} & 74.5\textsubscript{\textcolor{red!70!black}{↓11.8}} \\
        % Qwen-3-8B & 80.7 & 78.2 \\
        QwQ-32B & 80.7 & 78.2 \\
        Qwen-2.5-32B & {71.5} & {71.8} \\  \midrule
        
        \multicolumn{3}{c}{\cellcolor{lightgray!40}{Point-wise Judgment}} \\ \midrule
        Gemini-2.5-pro & 85.8\textsubscript{\textcolor{red!70!black}{↓0.2}} & \textbf{88.7}\textsubscript{\textcolor{green!70!black}{↑2.4}} \\
        Qwen-3-32B & 82.6\textsubscript{\textcolor{green!70!black}{↑6.1}} & 83.5\textsubscript{\textcolor{green!70!black}{↑9.0}} \\
        QwQ-32B & 83.8\textsubscript{\textcolor{green!70!black}{↑3.1}} & 83.9\textsubscript{\textcolor{green!70!black}{↑5.7}} \\
        Qwen-2.5-32B & 74.5\textsubscript{\textcolor{green!70!black}{↑3.0}} & 75.8\textsubscript{\textcolor{green!70!black}{↑4.0}} \\
        \bottomrule
    \end{tabular}
    \caption{Evaluation of reward reliability on hard and soft constraints of CFBench.
    Results show that: (1) LLM-based rewards achieve significantly lower reward precision than the rule checker; 
    and (2) point-wise judgment consistently outperforms batch judgment across most LLM-as-a-judge models.
    }
    \label{tab:llm_reward}
\end{table}

\subsection{Quantifying Reward Reliability}
% To this end, 
We evaluate the precision of reward signals from the reward model.
Concretely, we take the responses generated by the base model on CFBench, and evaluate them as follows:
(1) for responses involving hard constraints, we assess them using both a rule-based checker and an LLM-based reward model, respectively;
(2) for responses involving soft constraints, we evaluate them solely with the LLM-based reward model.
For comparison, we also test the frontier Gemini-2.5-Pro.
We establish ground-truth labels by human experts, details in Appendix \ref{sec:human_label}.
This allows us to quantify how well the reward models align with human evaluations.
Besides, we also compare point-wise judgment (assessing constraints individually) as an alternative to the conventional batch judgment (assessing all constraints simultaneously) for an instruction.
Notably, pointwise judgment is more time-consuming than batch judgment.

For hard constraints, rule-based verifiers act as a near-perfect ground truth.
In contrast, soft constraints rely on LLM-as-a-judge mechanisms, and we evaluate their agreement with the ground-truth labels.
As shown in Table \ref{tab:llm_reward}, we observe two key findings:
(1) Although stronger LLMs achieve higher alignment with humans, they significantly underperform compared to the rule checker,
e.g., the Qwen-3-32B lags behind by 19.5\% compared to rule checker.
This suggests that current LLM-based rewards suffer from a severe low reward-precision problem, which hampers the policy model's generalization in unverifiable constraint scenarios.
(2) The \textit{point-wise} judgment consistently outperforms batch judgment across open-sourced models, e.g., +6.1\% for Qwen-3-32B on hard constraints.
This suggests that when LLMs assess multiple constraints simultaneously, some form of bias interferes with reward precision. 

\begin{figure}[t]
    \small
    \centering
    \includegraphics[width=0.85\linewidth]{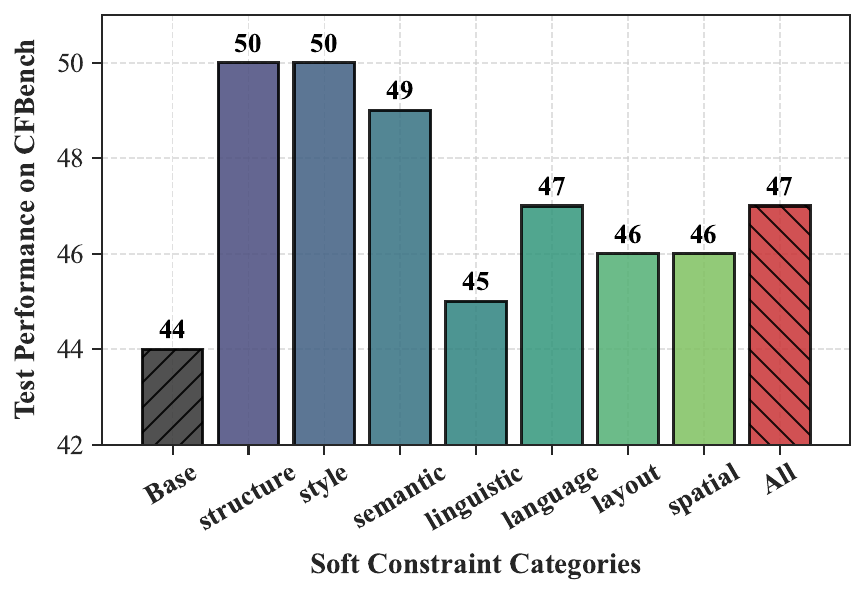}
  \caption{
    The generalization test performance of each soft constraint categories, on CFBench benchmarks. 
    It shows that \textit{style, structure}, and \textit{semantic} exhibit higher performance compared to other categories.
  }
  \label{fig:pre_study2}
\end{figure}

\subsection{Generalization Analysis of Constraints}
\label{sec:general}
To prevent bias of LLM-based rewards and gain insights for subsequent solutions,  
we analyze the generalization performance across all unverifiable soft constraint types.  
Specifically, to maximize coverage across all categories, we categorize soft constraints into seven categories: \textit{style, structure, semantic, linguistic, language, layout, and spatial}. 
Detailed definitions and statistical distribution are provided in Appendix~\ref{sec:soft_const}.

We split the training data into seven subsets and perform GRPO training on each subset separately.
We evaluate the test performance of models on CFBench benchmark (including all soft constraint categories and hard constraints), and show the results in Figure~\ref{fig:pre_study2}.
We observe that constraints in the categories of \textit{style, structure}, and \textit{semantic} exhibit higher generalization performance compared to other categories.
% For instance, category \textit{semantic} (ranked third) outperforms category \textit{language} (ranked fourth) by an average of 2\%.
These results suggest that training on these three constraint categories yields greater generalization gains.
We attribute this to the fact that these three constraints represent more foundational and generalized constraint patterns.
% We hypothesize that these three constraints encompass more fundamental and universal constraint patterns.

Overall, to improve reward precision, we can restrict LLMs to judge constraints one at a time, rather than assessing all constraints simultaneously.
By training only on constraints that support strong generalization, we can mitigate LLM reward bias while also improving judgment efficiency.
% Overall, to improve reward precision, we can restrict LLMs to judge the 受限的constraint types 逐个地, rather than all constraints simultaneously.
% by 训练 only on 约束 that support strong generalization，避免llm reward bais 的同时，提升的泛化性。
% It provides a key insight for the following solution.

\begin{figure*}[t]
    \centering
  \includegraphics[width=0.9\linewidth]{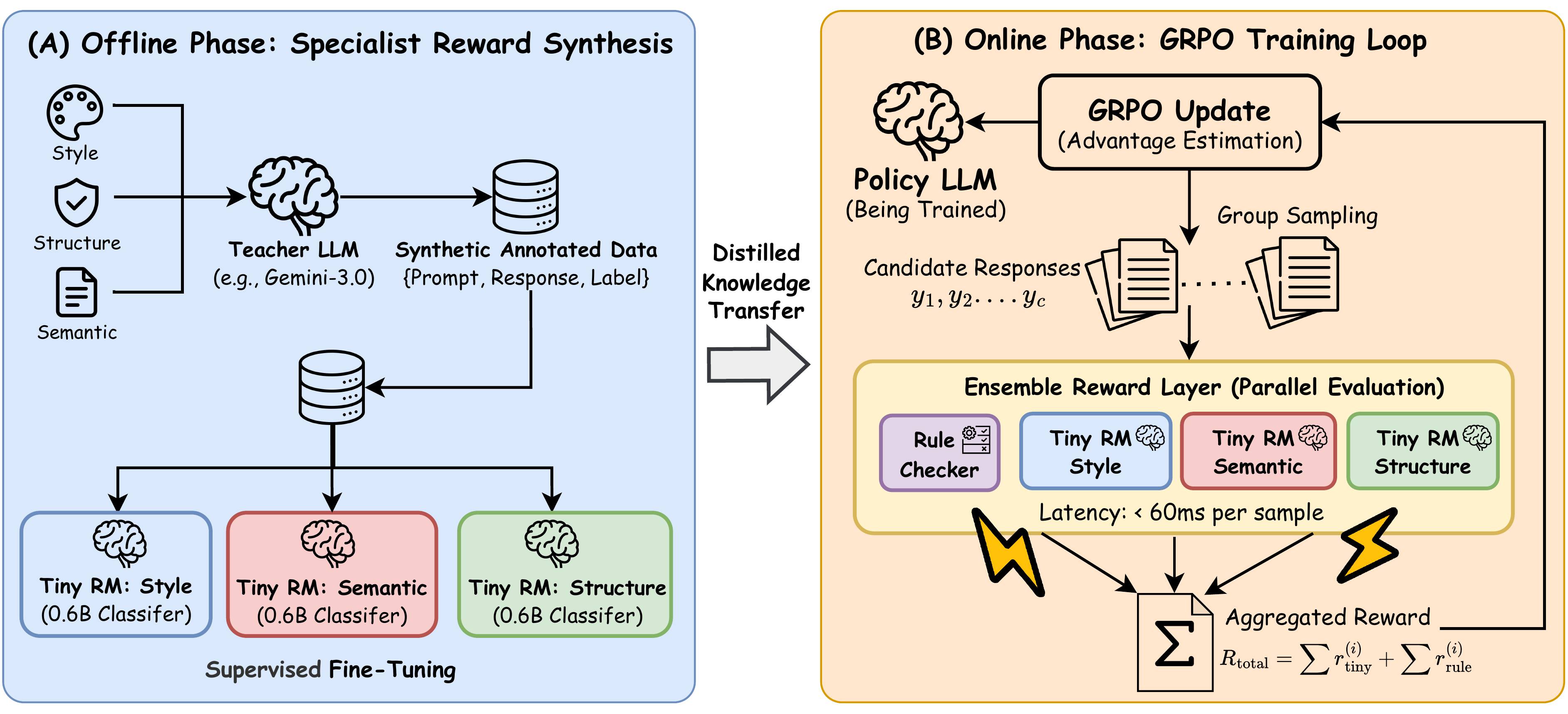}
  \caption{The constraint alignment framework (\textbf{TinyJudge}) via distilled specialist reward ensembles.
    It operates in two phases: 
    (A) Specialist Distillation: we respectively distill three soft constraint types from a teacher LLM into a suite of lightweight, high-precision classifiers. 
    (B) Accelerated GRPO Training: during RL training, these lightweight experts serve as a high-throughput ensemble reward model. 
    They provide multi-constraint signals for a sample in milliseconds, enabling efficient alignment with unverifiable constraints.}
  \label{fig:method}
\end{figure*}

\section{Methods}
This section introduces \textbf{TinyJudge}. 
Inspired by the above analysis, % in Section \ref{sec:general}
we incorporate only three high-generalization constraints alongside hard constraints.
We fine-tune three tiny models to provide precise reward signals in the GRPO loop, as shown in the framework in Figure \ref{fig:method}.

\subsection{Specialist Reward Synthesis}
In this offline phase, soft-constraint judgment knowledge is distilled from a powerful teacher model into a suite of lightweight specialist classifiers.
This process ensures that the resulting reward signal is both highly precise and computationally efficient for subsequent GRPO training.

To fully cover the three soft constraint types, we first augment the raw query set $\mathcal{Q}$ (Section \ref{sec:traindata}). 
For each query $q \in \mathcal{Q}$, we employ Gemini-3.0-Pro\footnote{https://gemini.google.com/} to synthesize and integrate a latent soft constraint $c_{soft}$ into the query to get the full instruction $I$.
Subsequently, we generate a diverse response pool $\mathcal{Y}$ by sampling from a heterogeneous set of models, including Qwen2.5-7B/32B-Instruct~\citep{qwen2025qwen25} and Llama3.2-3B~\citep{dubey2024llama}. 
This heterogeneity ensures that the specialists are exposed to varying levels of reasoning quality and common failure modes in instruction following.

Finally, for each triplet $(q, c_{soft}, y)$, where $y \in \mathcal{Y}$, we utilize Gemini-3.0-Pro to provide a binary assessment $r$ regarding the adherence to constraints.
we then fine-tune Qwen3-0.6B\footnote{To accelerate judgment, we disable the "thinking" mode.} as specialist judges.
A suite of other lightweight models is also evaluated in Section~\ref{sec:exp}.
Let $\mathcal{M}_k$ denote the $k$-th specialist model; 
we optimize it using a supervised fine-tuning objective:
\begin{equation}
    \small
    \mathcal{L}(\theta_k) = - \mathbb{E}_{(I, y, r) \sim \mathcal{D}_k} \sum_{t} \log P(r | I, y; \theta_k),
\end{equation}
where $\mathcal{D}_k$ is the specialist-specific dataset tailored to a particular constraint type (e.g., style or tone).

\subsection{Reward Engineering for GRPO Loop}
% as an efficient reward ensemble. 
In this online phase, TinyJudge computes a hybrid reward for each candidate response, enabling the model to align with both hard and soft constraints simultaneously.

% 平均集成 binary rule-based check for hard constraints  and specialist judges reward。
Concretely, once the specialist judges $\mathcal{M}_k$ are distilled, they are integrated into the GRPO training loop.
The total reward $R_{total}$ is synthesized through an additive ensemble of rule-based logic and neural specialists:
\begin{equation}
\small
R_{total}(q, r_i) = \frac{1}{N} \sum_{n=1}^{N}\mathcal{R}_{rule}^n(I, y_i) + \frac{1}{M} \sum_{k=1}^{M}\mathcal{M}_k(I, y_i),
\end{equation}
where $\mathcal{R}_{rule}$ and $\mathcal{M}$ both are the binary evaluators for hard and soft constraints, respectively. 
The GRPO objective is optimized to achieve a perfect score in $\mathcal{R}_{total}$.

By utilizing tiny models as specialist judges and performing inference in parallel with the policy rollout,
it can process a response in 10 milliseconds, effectively eliminating the latency bottleneck of LLM-based reward during the RL loop.

\section{Experiments}
\label{sec:exp}
% This section presents extensive experiments that validate the effectiveness of our proposed approach.

\subsection{Experimental Setup}
\label{sec:setup}
\textbf{Evaluation benchmarks}. 
We evaluate the models on five representative IF benchmarks, including IFEval~\citep{zhou2023instruction}, Multi-IF~\citep{he2024multi}, and IFBench~\citep{pyatkin2025generalizing}, all of which focus exclusively on verifiable hard constraints.
FollowBench~\citep{jiang2024followbench} and CFBench~\citep{zhang2025cfbench} cover a comprehensive range of mixed constraint types, including soft constraints.
The benchmark details are provided in Appendix \S\ref{sec:benchmark}.
For evaluation, we use the Instruction Satisfaction Rate (ISR; \citealp{zhang2025cfbench}) as the metric, which measures the model’s ability to satisfy all constraints within a given instruction.

\begin{table*}[t]
    \centering
    \small
    \begin{tabular}{l|ccc|cc|c}
        \toprule
         Model & \ding{168}IFEval & \ding{168}Multi-IF & \ding{168}IFBench & \ding{168}CFBench & \ding{171}FollowBench & {Average} \\
        \midrule 
        Gemini-3.0-Pro & 95.56 & 83.10 & 68.70 & 70.00 & 84.00 & 80.27 \\ 
        Claude-Sonnet-4.5 & 91.13 & 81.57 & 44.55 & 63.00 & 79.46 & 71.94 \\
        Deeepseek-V3.2 & 91.87 & 74.14 & 45.91 & 67.00 & 79.76 & 71.74 \\
        Qwen3-8B & 85.77 & 70.36 & 24.48 & 55.00 & 67.28 & 60.58 \\
        Qwen3-32B & 87.06 & 70.70 & 25.17 & 64.00 & 69.61 & 63.31 \\
        \midrule
        IF-RLVR\textsuperscript{$\dagger$} & 87.80 & --- & 53.70 & --- & --- & --- \\
        RECAST$^\dagger$  & 74.01 & --- & --- & --- & 63.23 & --- \\
        Qwen-IF$^\dagger$ & 78.90 & --- & --- & 52.00 & 63.80 & --- \\ 
        \midrule \rowcolor{lightgray!40}
        Qwen2.5-7B-inst & 72.46 & 51.05 & 28.91 & 44.00 & 61.40 & 51.56 \\ 
        \hspace{0.2cm}+ \textit{Qwen3-8B}\textsuperscript{$\ddagger$} & 77.82 & 57.15 & 29.59 & 48.00 & 68.64 & 56.24 \\  
        \hspace{0.2cm}+ \textit{Qwen2.5-32B}\textsuperscript{$\ddagger$} &  78.19 & 57.50 & 31.63 & 48.00 & 69.88 & 57.04 \\ 
        \hspace{0.2cm}+ \textit{Qwen3-32B}\textsuperscript{$\ddagger$} & 79.48 & 57.08 & 30.95 & 49.00 & 69.74 & 57.25 \\  
        % \hspace{0.2cm}+ QwQ-32B reward & 72.46 & 86.46 & 67.44 & 7.62 & 53.69 & -- \\  
        % \rowcolor{green!20} \hspace{0.2cm}+ TinyJudge-7B\textsuperscript{$\ddagger$} & 82.62\up{10.16} & 64.03\up{12.98} & 35.71\up{6.80} & 55.00\up{11.00} & 70.93\up{9.53} & 61.65\up{10.09} \\
        % \midrule \rowcolor{lightgray!40}
        \rowcolor{green!20} \hspace{0.2cm}+ TinyJudge-7B\textsuperscript{$\ddagger$} & 82.81\up{10.35} & 64.90\up{13.85} & 35.03\up{6.12} & 54.00\up{10.00} & 70.88\up{9.48} & 61.52\up{9.96} \\
        \midrule \rowcolor{lightgray!40}
        Qwen2.5-32B-inst & 81.70 & 64.45 & 33.67 & 57.00 & 73.06 & 61.98 \\ 
            \hspace{0.2cm}+ \textit{Qwen2.5-32B}\textsuperscript{$\ddagger$} & 83.55 & 68.23 & 33.67 & 58.00 & 75.20 & 63.73 \\ 
            \hspace{0.2cm}+ \textit{Qwen3-32B}\textsuperscript{$\ddagger$} & 84.47 & 68.29 & 35.71 & 60.00 & 74.35 & 64.56 \\  
            % \hspace{0.2cm}+ QwQ-32B reward & 72.46 & 86.46 & 67.44 & 7.62 & 53.69 & -- \\  
        \rowcolor{green!20}  \hspace{0.2cm}+ TinyJudge-32B\textsuperscript{$\ddagger$} & 86.51\up{4.81} & 73.57\up{9.12} & 41.83\up{8.16} & 64.00\up{7.00} & 77.01\up{3.95} & 68.58\up{6.60} \\
        \bottomrule
    \end{tabular}
    \caption{Evaluation results on the five IF benchmarks. 
    % The best result in each column is highlighted in bold, while the second-highest result is indicated with an underline formatting. 
    \ding{168} and \ding{171} denote hard-only constraints and mixed constraints (including soft constraints) benchmark, respectively.
    For RLVR-trained baselines$^\dagger$, we report only previously published results because the model checkpoints are not publicly available. 
    Models marked with the $\ddagger$ superscript are used as reward models.
    }
    \label{tab:final_exp}
\end{table*}

\textbf{Baseline.}
For our baseline comparisons, we report the performance of leading LLMs.
Additionally, we evaluate three recently RLVR-trained models derived from the same Base model:
RECAST\citep{guo2025recast} (verifiable data synthesis), IF-RLVR\citep{pyatkin2025generalizing} (multi-hard-constraint RL), and Qwen-IF\citep{ren2025instructions} (label-free self-supervised reward from instructions).  
In the setup where the LLM is used directly as a reward model\textsuperscript{$\ddagger$}, we train with a mix of three soft constraints alongside a hard constraint, aligning with our proposed method.
% All aim to improve instruction following through distinct RLVR strategies.
% \textbf{Models and Training.} 
% We select Qwen2.5-7B-Instruct as our base model and Qwen3-0.6B as the tiny model.
% The mixed hard and soft constraint data (details in Section \ref{sec:traindata}) is used for RL training.
% We employ standard GRPO~\citep{guo2025deepseek} algorithms for RLVR training from base model. 
More setup details shown in Appendix \S\ref{sec:exp_details}.

\subsection{Main Results}
\textbf{Overall Performance.}
The results are shown in Table~\ref{tab:final_exp}.
We observe that TinyJudge consistently outperforms the baselines,
e.g., achieving an average improvement of $\sim$ 10\% for the 7B model and 6.6\% for the 32B model.
Furthermore, consistent performance across benchmarks shows that limiting training to three soft constraint types still yields robust behavior.
Moreover, TinyJudge-32B (fast-think) outperforms the slow-thinking model Qwen3-32B by 5.47\%.
These results suggest that fine-tuning tiny models to judge constraints individually is an effective approach to enhancing IF capabilities.
The model trained with these high precision rewards can serve as a competitive alternative to more advanced large-scale LLMs.
To ensure the validity of the above findings, we scale our experiments across different model sizes (7B, 32B). 
As shown in Table~\ref{tab:final_exp}, these findings are consistently observed, suggesting that TinyJudge is robust across different base models.
Their training dynamics visualization is shown in Appendix~\ref{sec:visal}.

\subsection{Analysis Experiments}
Here, we present a detailed analysis, including ablation study on tiny models, training efficiency evaluation, and quantifying reward precision.

\subsubsection{Ablation Study on Tiny Model}
To investigate the impact of different tiny model configurations on reward modeling, we conduct an ablation study across three distinct settings.
For all experiments, we employ Qwen2.5-7B-Instruct as the policy model in RLVR. 
We evaluate a range of tiny models, including the Qwen2.5 and Qwen3 series, with scales varying from 0.5B to 4B parameter size. 
The configurations are defined as follows:

\begin{itemize}
    \item {Three Tiny Models for Three (\textit{3-for-3}, our):} Following the \textit{TinyJudge} approach, we employ three separate tiny models, each specialized for one of the three soft constraint types.
    \item {One Tiny Model for Three (\textit{1-for-3}):} A single tiny model is trained on the combined data of the three specific soft constraint types.
    \item {One Tiny Model for All (\textit{1-for-all}):} A single tiny model is trained on a mixture of all constraint data (i.e., the original \textit{mix} set).
\end{itemize}

\begin{table}[th]
    \small
    \centering
    \resizebox{\linewidth}{!}{
        \begin{tabular}{l|ccc}
            \toprule
            {\textbf{Tiny Model}} & \textbf{\ding{168}IFEval} & \textbf{\ding{168}CFBench} & \textbf{Avg.} \\
            \midrule
            Base & 72.46 & 44.00 & 58.23 \\
            \midrule
            \rowcolor{gray!20} \multicolumn{4}{c}{Three Tiny Models for Three (Our)} \\
            \midrule
            Qwen3-0.6B & \underline{82.81} & \underline{54.00} & \underline{68.41} \\
            Qwen3-1.7B & 81.70 & 53.00 & 67.35 \\
            Qwen3-4B-inst & 82.62 & \textbf{55.00} & \textbf{68.81} \\
            Qwen2.5-0.5B-inst & \textbf{83.36} & 52.00 & 67.68 \\
            Qwen2.5-1.5B-inst & 82.44 & 53.00 & 67.72 \\
            \midrule
            \rowcolor{gray!20} \multicolumn{4}{c}{One Tiny Model for Three} \\
            \midrule
            Qwen3-1.7B & 80.07 & 51.00 & 65.54 \\
            Qwen3-4B-inst & 80.62 & {53.00} & {66.81} \\
            \midrule
            \rowcolor{gray!20} \multicolumn{4}{c}{One Tiny Model for All} \\
            \midrule
            Qwen3-1.7B & 78.74 & 46.00 & 62.37 \\
            Qwen3-4B-inst & 81.70 & 46.00 & 63.85 \\
            \bottomrule
        \end{tabular}
    }
    \caption{Ablation study on tiny models. 
    The best result in each column is highlighted in bold, while the second-highest result is indicated with an underline formatting.}
    \label{tab:tinymodel}
\end{table}

Results presented in Table \ref{tab:tinymodel} reveal several insights. 
Within the \textit{3-for-3} setting, Qwen3 models exhibit consistent trends across different sizes (e.g., Qwen3-4B-Inst and Qwen3-0.6B differ by only 0.4 points).
Notably, Qwen3 architectures exhibit better reward modeling performance than Qwen2.5 at the same model scale. 
Furthermore, we observe that the \textit{3-for-3} configuration consistently yields superior results compared to the \textit{1-for-all} and \textit{1-for-3} settings using the same tiny model backbone. 
For instance, when using Qwen3-4B-Inst as the reward model, the \textit{3-for-3} setting outperforms \textit{1-for-all} and \textit{1-for-3} by 2\% and 5\%, respectively.
This evidence strongly supports the efficacy of our proposed architectural design for reward models.

\subsubsection{Training Efficiency}
\begin{figure}[th]
    \centering
  \includegraphics[width=\columnwidth]{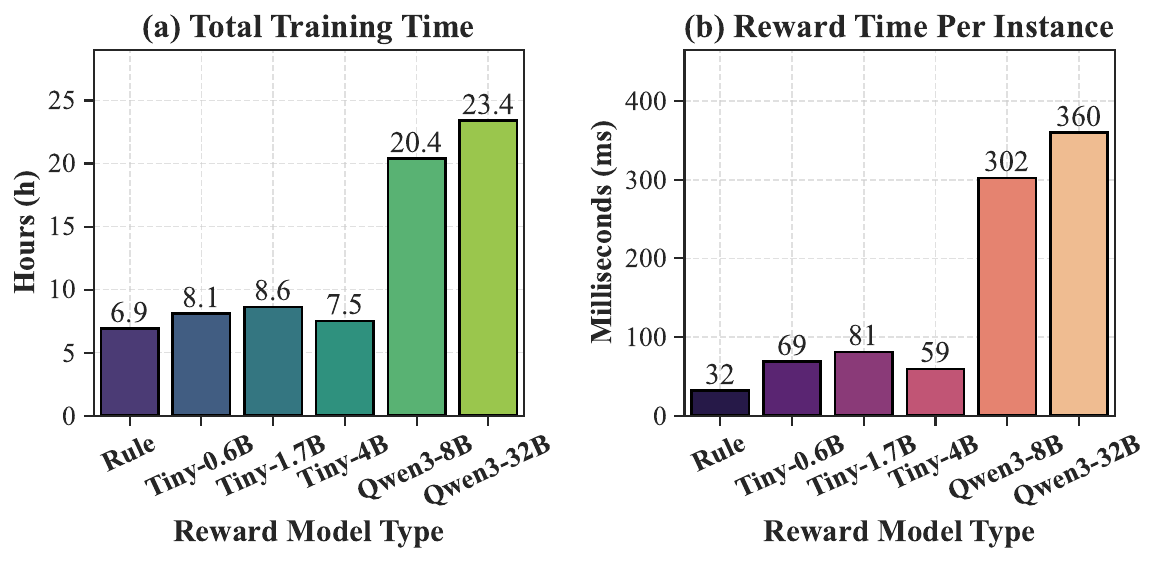}
  \caption{Comparison of training efficiency. 
  Our tiny reward strategy significantly reduces computational cost compared to LLM-based rewards, achieving efficiency close to that of rule-based rewards.
    }
  \label{fig:time_cost2}
\end{figure}

To assess training efficiency, we measured the computational overhead under the same setup as Section~\ref{sec:traindata}.
Figure \ref{fig:time_cost2} compares the Qwen3-Tiny reward model against traditional baselines in \textit{3-for-3} setup.
It reveals that our reward strategy reduces computational cost compared to direct LLM rewards by $\sim3\times$.
For example, Tiny-4B achieves a 312\% speedup over Qwen3-32B in total training time, 610\% for judging latency per response.
Remarkably, the strategy achieves a processing speed comparable to that of rule-based rewards.
These results demonstrate that our proposed strategy effectively mitigates the time overhead arising from LLM rewards, accelerating the training process.

\begin{figure}[th]
    \centering
  \includegraphics[width=0.80\columnwidth]{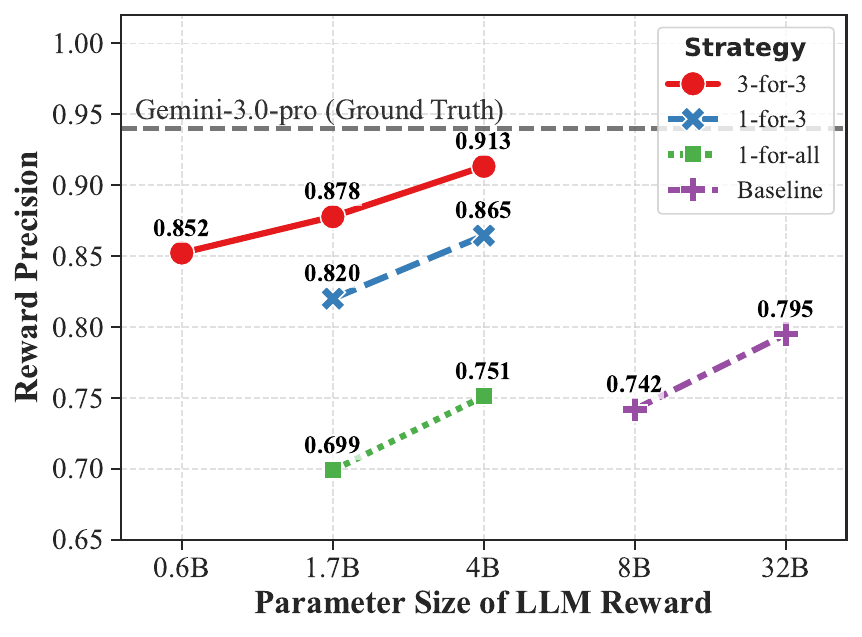}
  \caption{Quantifying reward precision in three soft constraint types across various reward models. 
  Our proposed \textit{3-for-3} reward strategy consistently achieves high reward precision.
    }
  \label{fig:llm_reward_prec}
\end{figure}

\subsubsection{Quantifying Reward Precision}
We quantify reward precision to verify that our proposed TinyJudge effectively enhances reward reliability. 
We utilize a 5\% held-out validation set and establish ground-truth labels by averaging five independent evaluations from Gemini-3.0-Pro.
We then measure the alignment between various reward models (including Qwen3-Tiny configurations in \textit{3-for-3}, \textit{1-for-all}, and \textit{1-for-3} setups) and the ground-truth.
As illustrated in Figure \ref{fig:llm_reward_prec}, the tiny model within the \textit{3-for-3} setup achieves the highest reward precision. 
For instance, the tiny 4B model outperforms Qwen3-32B by 12\% in reward precision.
This confirms that the TinyJudge strategy improves instruction-following generalization by directly improving the precision of reward signal.

\section{Conclusion}
In conclusion, this paper identifies and addresses the critical bottlenecks of reward bias and computational inefficiency in scaling soft constraints for RLVR.
We propose a novel framework that replaces heavy LLM judges with an ensemble of specialized, tiny expert models tailored for high-generalization semantic constraints.
It not only achieves a performance gain through higher reward precision but also delivers a 3$\times$ reduction in total training time.
These findings establish a highly efficient and robust paradigm for aligning models with complex, unverifiable human instructions without 
large-scale models.
% The related work is shown in Section \ref{sec:related_work}.

\section*{Limitations}
First, the selection and combination of constraints lack a systematic optimization framework. 
While we focus on high-generalization soft constraints, we have not yet established a theoretical or empirical boundary for the optimal number of expert models to be ensembled. 
Additionally, the reasons behind the strong generalization of these specific constraints warrant further investigation.
% Consequently, the framework may exhibit limited coverage for long-tail soft constraints that do not demonstrate strong generalization patterns, potentially leading to sub-optimal reward signals for highly specialized instructions.

Second, the performance of tiny reward models is inherently capped by the teacher LLM. 
Since our training process relies on distilling expertise from frontier models, any systematic biases or reward inaccuracies present in the teacher model will inevitably be inherited by the specialized tiny judges. 

Third, the \textit{3-for-3} architecture faces scalability challenges at scale.
Although the ensemble of tiny models offers high inference efficiency, a tiny model requires distillation and training whenever a new constraint type is introduced.

\section*{Acknowledgements}
The research in this article is supported by the New Generation Artificial Intelligence of China (2024YFE0203700), National Natural Science Foundation of China under Grants U22B2059 and 62576124.

% \section*{Acknowledgments}

\bibliography{anthology}

\appendix

\section{Contribution}
In summary, our contributions are as follows:
\begin{itemize}
    \item Empirical Finding: We conduct a systematic analysis of LLM-as-a-judge in RLVR, revealing that existing paradigms suffer from severe leniency bias in multi-constraint scenarios and higher computational costs.
    \item Methodological Innovation: We propose \textbf{TinyJudge}, a scalable framework that distills reward signals into tiny, specialized expert models (e.g., 0.6B) based on constraint generalization analysis. 
    This enables precise feedback for unverifiable soft constraints without relying on heavy online judges.
    \item Good Results: Extensive evaluations across five benchmarks show that TinyJudge achieves a $\sim$10\% performance gain over baselines. Crucially, it improves reward precision by 12\% and reduces training time by 3$\times$, matching the efficiency of rule-based methods.
\end{itemize}

\section{Details in Reward Reliability}
\label{sec:human_label}

To establish a rigorous ground-truth reference for validating reward signal reliability, we implemented a multi-stage human evaluation protocol. 

\paragraph{Data Sampling.}
We conducted stratified random sampling from the CFBench validation set to ensure diverse coverage of difficulty levels and constraint types. Specifically, we sampled $N=100$ instances from the Hard-Only subset (e.g., formatting, symbol restrictions) and $N=100$ instances from the Soft-Only subset (e.g., stylistic tone, creative writing).

\paragraph{Expert Annotation Protocol.}
The annotation task was assigned to three independent human experts, all of whom are graduate-level researchers in NLP with high proficiency in English. To mitigate potential bias, we employed a \textbf{double-blind procedure}: annotators were blinded to the model sources and the automated reward scores. 
We developed a strict \textit{Adherence Rubric} prior to annotation:
\begin{itemize}
    \item \textbf{Hard Constraints:} Evaluated on a binary scale (Pass/Fail). A response is marked as specific `Fail' if it violates any strictly verifiable rule (e.g., missing a JSON key).
    \item \textbf{Soft Constraints:} Evaluated on a 5-point Likert scale regarding the degree of instruction following (5=Perfect Alignment, 1=Complete Violation). For binary classification analysis, scores $\geq 4$ are converted to `Pass'.
\end{itemize}

\paragraph{Quality Assurance and Adjudication.}
To ensure label consistency, we calculated Fleiss' Kappa ($\kappa$) to measure Inter-Annotator Agreement (IAA). We observed a substantial agreement for hard constraints ($\kappa=0.89$) and moderate-to-high agreement for soft constraints ($\kappa=0.72$).
For instances with significant disagreement (variance $>1.0$ on the Likert scale or split votes on binary labels), a fourth senior expert acted as an adjudicator to determine the final ground-truth label. The final reference score is derived from the adjudicated consensus rather than a simple average.

\section{Additional Experiments}
\label{sec:exp_details}
In this section, we report additional experimental results.

\subsection{Implementation Details}
\label{sec:grpo_details}
For reinforcement learning, we implemented GRPO based on the MindSpeed-RL\footnote{https://gitcode.com/Ascend/MindSpeed-RL} training framework. 
Each RL training run completed on a cluster of 64 Ascend 910b NPUs (configured as 8 nodes × 8 NPUs). 
The hyperparameters used are detailed in Table \ref{tab:grpo_config}.
\begin{table}[ht]
    \centering
    \small
    \begin{tabular}{l|c}
        \toprule
        \textbf{Hyperparameter} & \textbf{Value} \\
        \midrule
        \multicolumn{2}{c}{\textit{Data Configuration}} \\
        \midrule
        Global Batch Size & 128 \\
        Max Prompt Length & 2048 \\
        Max Response Length & 6144 \\
        Micro Batch Size & 4 \\
        Train Steps & 400 \\
        \midrule
        \multicolumn{2}{c}{\textit{Rollout Configuration}} \\
        \midrule
        Rollout Name & vllm \\
        GPU Memory Utilization & 0.6 \\
        Number of Rollouts & 8 \\
        Temperature & 1.0 \\
        Tensor Model Parallel Size & 1 \\
        Top\_P & 1.0 \\
        \midrule
        \multicolumn{2}{c}{\textit{RL Optimization}} \\
        \midrule
        Learning Rate & 1e-6 \\
        LR Decay Style & constant \\
        Mini Batch Size & 128 \\
        KL Loss & 0.001 \\
        \bottomrule
    \end{tabular}
    \caption{The configurations for RLVR training.}
    \label{tab:grpo_config}
\end{table}

\subsection{Robustness Check}
The result is shown in Table~\ref{tab:robust}.

\begin{table*}[th]
    \centering
    \small
    \begin{tabular}{l | ccc|cc|c}
        \toprule
         Model & \ding{168}IFEval & \ding{168}Multi-IF & \ding{168}IFBench & \ding{171}CFBench & \ding{171}FollowBench & \textbf{Average} \\
        \midrule \rowcolor{lightgray!60}
        Qwen2.5-7B-inst & 72.46 & 51.05 & 28.91 & 44.00 & 61.40 & 51.56 \\ 
        \hspace{0.2cm} w/ \textit{hard-only} & 80.78 & 58.89 & 31.63 & 49.00 & 68.96 & \textbf{57.85} \\
        \hspace{0.2cm} w/ \textit{soft-only} & 77.82 & 54.44 & 27.89 & 47.00 & 68.73 & \underline{55.18} \\
        \hspace{0.2cm} w/ \textit{mix} & 78.37 & 58.25 & 29.59 & 51.00 & 69.07 & 57.26 \\
        \midrule \rowcolor{lightgray!60}
        Qwen2.5-32B-inst & 81.70 & 64.45 & 33.67 & 57.00 & 73.06 & 61.98 \\ 
        \hspace{0.2cm} w/ \textit{hard-only} & 84.10 & 68.59 & 35.71 & 60.00 & 75.12 & 64.70 \\
        \hspace{0.2cm} w/ \textit{soft-only} & 82.99 & 66.04 & 30.95 & 58.00 & 74.19 & \underline{62.43} \\
        \hspace{0.2cm} w/ \textit{mix} & 83.55 & 68.87 & 37.75 & 60.00 & 74.99 & \textbf{65.03} \\
        \midrule \rowcolor{lightgray!60}
        Qwen3-8B & 85.77 & 70.36 & 24.48 & 55.00 & 67.28 & 60.58 \\ 
        \hspace{0.2cm} w/ \textit{hard-only} & 88.54 & 73.37 & 25.17 & 57.00 & 67.79 & \textbf{62.37} \\
        \hspace{0.2cm} w/ \textit{soft-only} & 86.32 & 70.97 & 27.21 & 56.00 & 67.21 & \underline{61.54} \\
        \hspace{0.2cm} w/ \textit{mix} & 86.51 & 72.73 & 25.85 & 57.00 & 67.70 & 61.96 \\
        \midrule \rowcolor{lightgray!60}
        Llama3.2-3B-inst & 74.12 & 40.99 & 20.74 & 17.00 & 50.40 & 40.65 \\ 
        \hspace{0.2cm} w/ \textit{hard-only} & 78.00 & 53.00 & 25.85 & 22.00 & 53.65 & 46.50 \\
        \hspace{0.2cm} w/ \textit{soft-only} & 74.86 & 44.75 & 24.82 & 22.00 & 54.11 & \underline{44.11} \\
        \hspace{0.2cm} w/ \textit{mix} & 79.30 & 51.31 & 24.48 & 25.00 & 53.61 & \textbf{46.74} \\
        \bottomrule
    \end{tabular}
    \caption{Evaluation results on five benchmarks for robustness checking. 
    We use Qwen3-32B as the reward model for all \textit{soft-only} and \textit{mix} settings.
    \ding{168} and \ding{171} denote hard-only constraints and mixed constraints (including soft constraints) benchmarks, respectively.
    Bold and underlined values indicate the \textbf{best} and \underline{worst} performances, respectively.
    The \textit{hard-only} models consistently outperform \textit{soft-only} models across various base model configurations, even on OOD mixed constraints benchmarks.
    }
    \label{tab:robust}
\end{table*}

\subsection{Training Dynamics Visualization.}
\label{sec:visal}
\begin{figure*}[th]
    \small
    \centering
      \begin{subfigure}[t]{0.33\linewidth}
        \centering
        \includegraphics[width=0.99\linewidth]{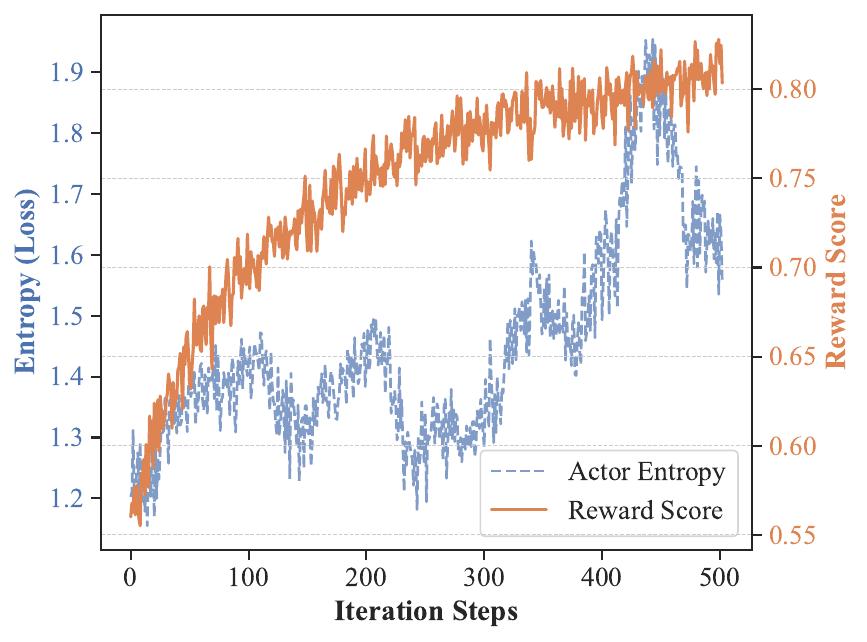} 
        \caption{Reward with Qwen3-0.6B in \textit{3-for-3}.}
        \label{fig:exp2a}
    \end{subfigure}
    \hfill
      \begin{subfigure}[t]{0.33\linewidth}
        \centering
        \includegraphics[width=0.99\linewidth]{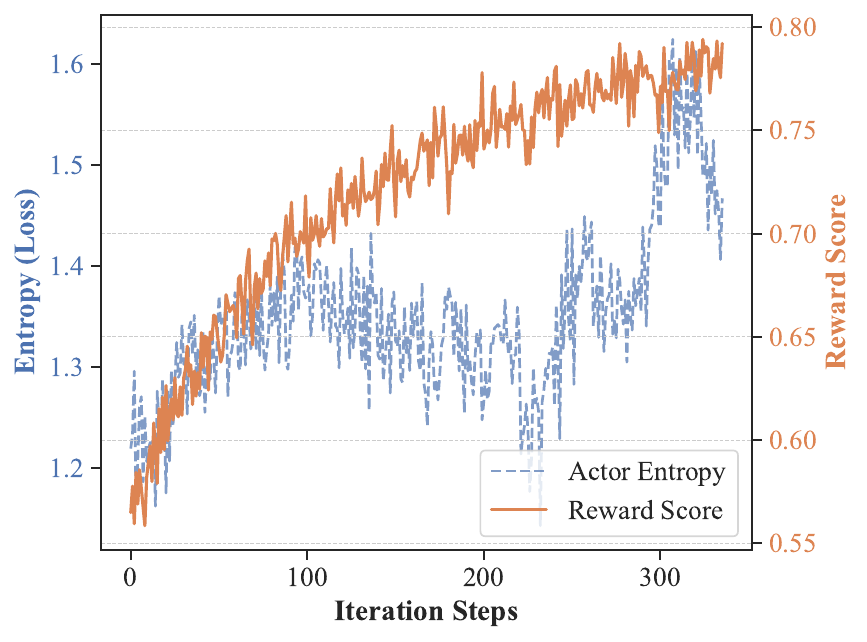} 
        \caption{Reward with Qwen3-1.7B in \textit{3-for-3}.}
        \label{fig:exp2b}
    \end{subfigure}
      \begin{subfigure}[t]{0.33\linewidth}
        \centering
        \includegraphics[width=0.99\linewidth]{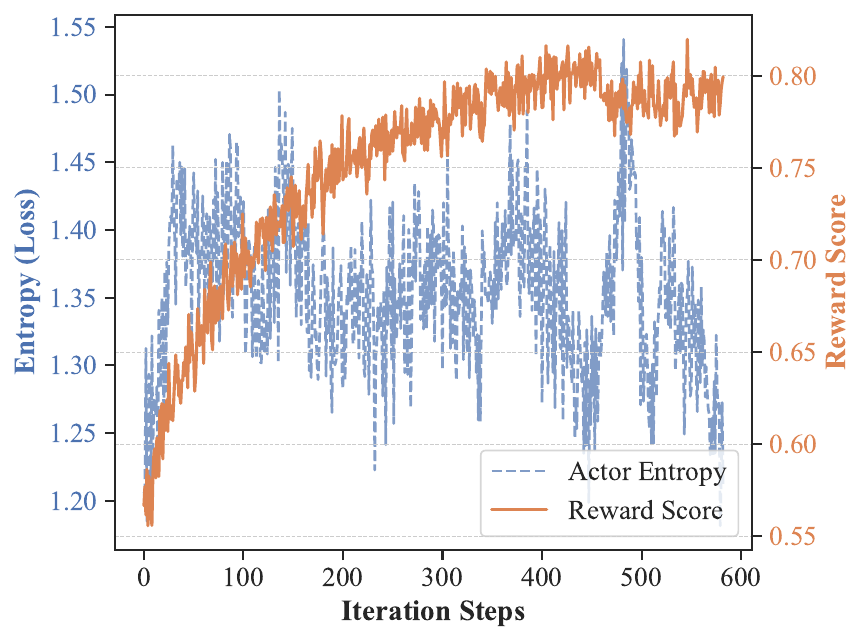} 
        \caption{Reward with Qwen3-4B in \textit{3-for-3}.}
        \label{fig:exp2c}
    \end{subfigure}
        \\
    \begin{subfigure}[t]{0.33\linewidth}
        \centering
        \includegraphics[width=0.99\linewidth]{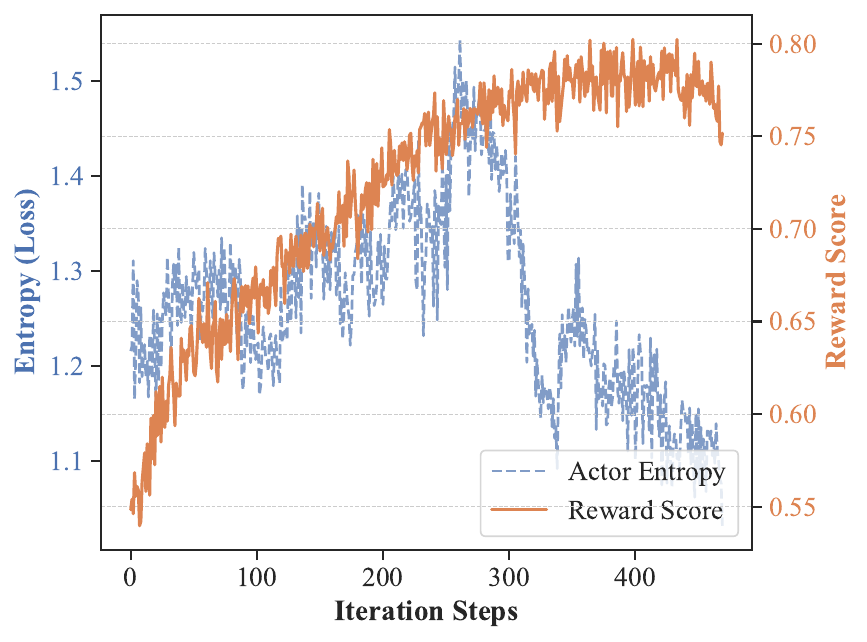} 
        \caption{Reward with Qwen2.5-0.5B in \textit{3-for-3}.}
        \label{fig:exp4a}
    \end{subfigure}
    % \hfill
      \begin{subfigure}[t]{0.33\linewidth}
        \centering
        \includegraphics[width=0.99\linewidth]{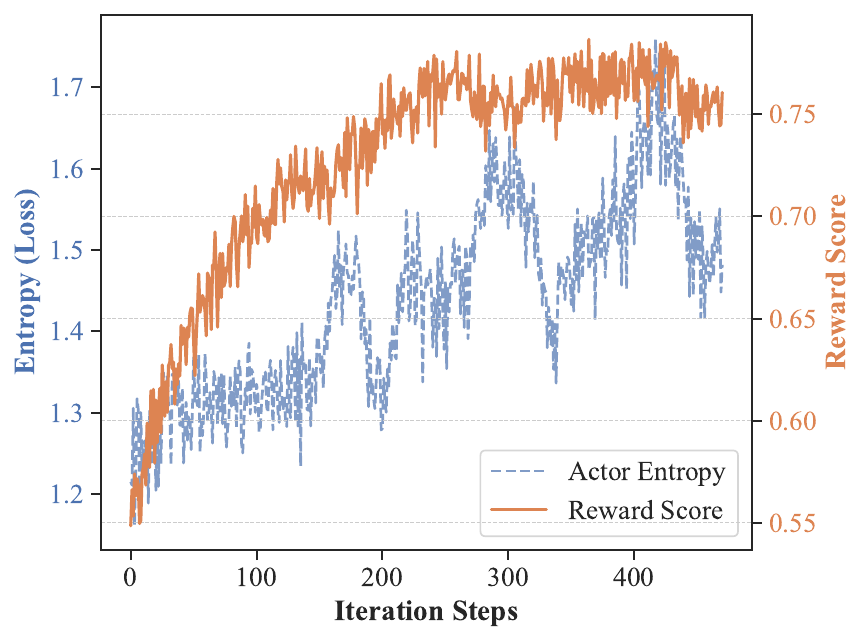} 
        \caption{Reward with Qwen2.5-1.5B in \textit{3-for-3}.}
        \label{fig:exp4b}
    \end{subfigure}
    % \\
    % \begin{subfigure}[t]{0.33\linewidth}
    %     \centering
    %     \includegraphics[width=0.99\linewidth]{./reward_qwen3_1.7tiny_1_for_3.pdf} 
    %     \caption{Reward with Qwen3-1.7B in \textit{1-for-3}.}
    %     \label{fig:exp3a}
    % \end{subfigure}
    % % \hfill
    %   \begin{subfigure}[t]{0.33\linewidth}
    %     \centering
    %     \includegraphics[width=0.99\linewidth]{./reward_qwen3_4tiny_1_for_3.pdf} 
    %     \caption{Reward with Qwen3-4B in \textit{1-for-3}.}
    %     \label{fig:exp3b}
    % \end{subfigure}
  \caption{Visualization of training dynamics for various tiny models.  }
  \label{fig:visualize}
\end{figure*}

We visualize the various model training curves in Figure~\ref{fig:visualize}.
% It includes the \textit{3-for-3}, \textit{1-for-all}, and \textit{1-for-3} configurations using various tiny models, as shown in Table~\ref{tab:tinymodel}.
In this paper, we use the checkpoint at step 400 as the final model.

\subsection{Impact on General Capabilities and Robustness}
\label{appendix:general_eval}

To investigate whether fine-tuning on strict instruction-following (IF) constraints introduces negative transfer or compromises the model's broader utility, we evaluate the TinyJudge series across three representative benchmarks: \textbf{GSM8K} for mathematical reasoning, \textbf{MMLU} for world knowledge, and \textbf{WritingBench} for open-ended generation quality.

\paragraph{Performance Comparison}
As summarized in Table~\ref{tab:general_capabilities}, the integration of TinyJudge does not degrade general capabilities; instead, it yields consistent improvements, particularly in larger model scales.

\begin{table}[th]
    \centering
    \small
    \begin{tabular}{lccc}
        \toprule
        \textbf{Model} & \textbf{GSM8K} & \textbf{MMLU} & \textbf{WritingBench} \\ \midrule
        Qwen-7B-inst & 92.10\% & 74.54\% & 5.995 \\
        \textbf{TinyJudge-7B} & 91.88\% & 73.20\% & 5.972 \\ \midrule
        Qwen-32B-inst & 95.44\% & 78.02\% & 6.116 \\
        \textbf{TinyJudge-32B} & \textbf{95.60\%} & \textbf{80.23\%} & \textbf{6.392} \\ 
        \bottomrule
    \end{tabular}
    \caption{Comparison of general capabilities between TinyJudge and Qwen-Instruct baselines.}
    \label{tab:general_capabilities}
\end{table}

\paragraph{Analysis of Generalization and Failure Modes}
Our analysis reveals several key insights regarding the robustness of the TinyJudge framework:
\begin{itemize}
    \item \textbf{Synergistic Reasoning and Knowledge:} For the 32B model, TinyJudge achieves a 2.21\% improvement on MMLU and a marginal gain on GSM8K. This suggests that learning to strictly follow complex constraints enhances the model's overall logical coherence and retrieval precision, rather than acting as a restrictive bottleneck.
    \item \textbf{Stylistic Refinement:} The significant performance boost on WritingBench (6.392 vs. 6.116) indicates that fine-tuning for constraint adherence directly translates to better mastery of stylistic and structural nuances, leading to higher perceived output quality.
    \item \textbf{Absence of Negative Transfer:} The 7B variant remains highly competitive with its baseline, confirming that our tuning process does not induce common failure modes.
\end{itemize}

In conclusion, these results demonstrate that the TinyJudge approach enhances specific instruction-following capabilities while preserving, and in some cases bolstering, the model's general-purpose proficiency and robustness.

\section{Taxonomy of Soft Constraints}
\label{sec:soft_const}
For model training, we utilize the \textit{VerInstruct} dataset~\citep{peng2025verif}, comprising 22,000 instances. 
It includes two constraint types: {soft constraints} (77.7\%), which are evaluated via LLM-as-a-judge, and {hard constraints} (22.3\%).
We analyze the generalization performance across all unverifiable soft constraint types in this dataset.  
Specifically, to maximize coverage across all categories, we categorize soft constraints into seven types: \textit{style, structure, semantic, linguistic, language, layout, and spatial}. 
Their definition are shown in Table \ref{tab:definition}.
The category distribution of soft constraints is summarized in Table \ref{tab:dataset_distribution}.
% style, structure, content, linguistic, language, layout, and spatial

\begin{table*}[th]
    \centering
    \small 
    \begin{tabular}{c|p{6cm}p{7cm}}
    \toprule
    \textbf{Category} & \textbf{Definition} & \textbf{Example} \\ 
        \midrule
        Style & Specifies the overall atmosphere (vibe), narrative tone, or the persona (role) of the writer. & Answer in the persona of a weary film noir detective; the tone should be cold, cynical, and filled with a sense of fatalism. \\ \addlinespace
        
        Structure & Focuses on the abstract logical layout or imitation of specific non-fixed formats, emphasizing logical flow. & Adopt an `echoing' structure: ensure the final paragraph provides a specific response to the philosophical hypothesis in the first paragraph. \\ \addlinespace
        
        Semantic & Regulates the introduction or exclusion of specific semantic content, or the use of particular rhetorical devices. & Explain code refactoring using a `pruning a bonsai' metaphor, and strictly avoid mentioning any specific programming language names. \\ 
        
        \midrule \midrule
        Linguistic & Concerns atomic-level grammatical constructions, parts-of-speech restrictions, or morphological rules. & Do not use any passive voice constructions, and ensure that no single sentence exceeds a length of 15 words. \\ \addlinespace
        
        Language & Governs rules for switching between languages and language alignment within specific cultural contexts. & Respond in Chinese, but naturally embed authentic English technical terms when discussing system architecture. \\ \addlinespace
        
        Layout & Focuses on the visual distribution of text, paragraph density, and the specific formatting of lists. & The response must be divided into three paragraphs of equal length, and all list items must start with a custom symbol (e.g., $\diamond$). \\ \addlinespace
        
        Spatial & Mandates the absolute spatial position or visual weight of specific information within the text stream. & At the very end of the response, naturally embed a specific closing statement and bold every technical term mentioned. \\ \bottomrule
    \end{tabular}
    \caption{Definitions and Examples of Constraint Categories}
    \label{tab:definition}
\end{table*}

\begin{table*}[th]
    \centering
    % \small
    \begin{tabular}{c|lr}
        \toprule
        \textbf{Constraint Type} & \textbf{Original Data Keys} & \textbf{Count} \\ \midrule
        \textbf{Style} & Desired\_Writing\_Style & 15,343 \\
          \textbf{Structure} & Hierarchical\_Instructions & 16,456 \\
          \textbf{Semantic} & Semantic\_elements, Specific\_Literary\_Devices & 34,782 \\
        \textbf{Linguistic} & Specific\_Grammatical, Morphological & 36,162 \\
          \textbf{Language} & Multi-lingual\_Constraints & 17,185 \\
          \textbf{Layout} & Paragraphs\_Constraints, Item\_Listing\_Details & 24,868 \\
          \textbf{Spatial} & Specific\_Sentence, Special\_Format, Key\_Formatting & 17,340 \\ \bottomrule
    \end{tabular}
    \caption{Summary of soft constraint categories distribution}
    \label{tab:dataset_distribution}
\end{table*}

\section{Related Work}
\label{sec:related_work}
{Instruction Following} requires models to generate responses that satisfy user's complex instruction constraints.
% Adhering to complex constraints, particularly those involving a greater number and variety of conditions, remains a challenge for LLMs~\citep{zhang2025cfbench,tam2024let,sun2023evaluating}.
Earlier works favored sophisticated data synthesis, such as self-dialogue in AutoIF~\citep{dong2024self}, rule extraction in RNR~\citep{wang2024rnr}, or verifiable data generation in VFF~\citep{wang2025verifiable}, to scale supervised fine-tuning or DPO~\citep{kim2025systematic,jiang2024followbench,chung2024scaling,pyatkin2025generalizing}. 
Recently, the research community is actively advancing IF studies, evidenced by benchmarks such as FollowBench~\citep{jiang2024followbench} and CFBench~\citep{zhang2025cfbench}, shifting focus from hard-only constraints toward scaling diverse, structured constraints to improve generalization in complex real-world scenarios.
To achieve this goal, the RLVR training paradigm~\citep{peng2025verif} is a mainstream approach currently being actively explored.

RLVR has garnered attention for its ability to incentivize reasoning in LLMs through rule-based rewards~\citep{yue2025does,zhu2025surprising,dai2025s,2025kimik2}.
This paradigm has demonstrated remarkable efficacy in reasoning domains, e.g., math and coding~\citep{zeng2025simplerl,fatemi2025concise,liu2025deepseek}.
However, the application of RLVR remains largely restricted to tasks with explicit, verifiable ground truths.
To extend this success to soft constraints in IF, researchers have relied on LLM-as-a-judge to provide reward signals~\citep{pyatkin2025generalizing,guo2025recast,peng2025verif,lambert2024tulu}.
In contrast, our work identifies the LLM reward hacking bottlenecks.
To address it, we introduce expert-level tiny models designed to minimize training overhead and mitigate reward hacking. 

\section{Benchmark Details}
\label{sec:benchmark}
To comprehensively assess the instruction-following capabilities of LLMs, we utilize five distinct benchmarks. 
These datasets cover hard and soft constraints, and also assess generalization to unseen constraints, multi-level difficulty, and multi-turn multilingual interactions. 
Table~\ref{tab:benchmarks} summarizes the statistics of these datasets.

\textbf{IFEval} ~\cite{zhou2023instruction} is a widely adopted benchmark designed to evaluate the ability of LLMs to follow objective and verifiable instructions. It consists of around 500 prompts containing 25 types of verifiable constraints (e.g., word count limits, formatting requirements).

\textbf{Multi-IF}~\cite{he2024multi} extends the scope of instruction following to multi-turn and multilingual settings. It contains 4,501 samples spanning 8 languages. The dataset is constructed by expanding single-turn verifiable instructions into coherent three-turn dialogues. It serves as a stress test for maintaining instruction adherence over long contexts and across diverse linguistic distributions.

\textbf{IFBench}~\cite{pyatkin2025generalizing} addresses the issue of model overfitting to common instruction datasets. It focuses on evaluating the generalization capabilities of models by introducing 58 novel, unseen, and challenging verifiable constraints. It employs strict code-based verification modules to measure performance on out-of-domain instructions.

\textbf{FollowBench}~\cite{jiang2024followbench} evaluates the robustness of LLMs through a multi-level difficulty mechanism. It contains 820 instructions across five fine-grained categories (Content, Situation, Style, Format, Example). The benchmark is constructed by incrementally adding constraints to seed instructions, creating a difficulty gradient (Level 1 to Level $N$).

\textbf{CFBench}~\cite{zhang2025cfbench} is a benchmark comprising 1,000 high-quality samples derived from the real world. It features a hierarchical taxonomy with 10 major constraint categories (e.g., style, logical rules, numerical) and over 25 subcategories. CFBench is designed to simulate complex tasks, including both an \textit{Easy} and \textit{Hard} subset to test models on varying degrees of constraint complexity.

\begin{table*}[th]
    \centering
    \small
    \begin{tabular}{l|cccc}
        \toprule
        \textbf{Dataset} & \textbf{Size (Samples)} & \textbf{Constraint Types} & \textbf{Key Features} & \textbf{Eval. Method} \\
        \midrule
            {IFEval}~\cite{zhou2023instruction} & 500 & hard / 25 & Verifiable, Objective & Code-based \\
            {IFBench}~\cite{pyatkin2025generalizing} & 300 & hard / 58 & Unseen Constraint & Code-based \\
            {Multi-IF}~\cite{he2024multi} & 4,501 & hard / 25 & Multi-turn, Multilingual  & Code-based \\
            {FollowBench}~\cite{jiang2024followbench} & 820 & mixed / 5 & Multi-level Difficulty  & Code + LLM \\
            {CFBench}~\cite{zhang2025cfbench} & 1,000 & mixed / 25 & Complex Scenarios & LLM \\
        \bottomrule
    \end{tabular}
    \caption{Statistics and characteristics of the instruction following benchmarks.}
    \label{tab:benchmarks}
\end{table*}

\end{document}